\documentclass[letterpaper, 10 pt, conference]{ieeeconf}  % Comment this line out if you need a4paper
\IEEEoverridecommandlockouts                              % This command is only needed if 
\usepackage{times}
\usepackage{multicol}
\usepackage[bookmarks=true]{hyperref}
\usepackage{graphicx} % for pdf, bitmapped graphics files
\usepackage{amsmath} % assumes amsmath package installed
\usepackage{amssymb}  % assumes amsmath package installed
\usepackage{bm} % for using bold lowercase greek letters
\usepackage{array}
\usepackage{colortbl}	% to color table background
\usepackage[table]{xcolor}
\usepackage{subfig}  
\usepackage{algorithmicx,algpseudocode}
\usepackage{algorithm}
\usepackage{paralist}
\usepackage{multirow}
\usepackage{siunitx}           % for typesetting micrometer with an unright "mu"

\newcommand{\pder}[2]{\frac{\partial#1}{\partial#2}}		% partial derivative
\DeclareMathOperator*{\st}{s.t.}						% subject to
\DeclareMathOperator*{\half}{\frac{1}{2}}					% one half
\newcommand{\mat}[1]{\ensuremath{\begin{bmatrix}#1\end{bmatrix}}}	% matrix
\newcommand{\dx}[1]{\ensuremath{\delta x_{#1}}}					% dx
\newcommand{\du}[1]{\ensuremath{\delta u_{#1}}}					% du
\newcommand{\DX}[0]{\ensuremath{\Delta X}}						% DX
\newcommand{\DU}[0]{\ensuremath{\Delta U}}						% DU
\newcommand{\T}[0]{\ensuremath{\top}}							% transpose symbol
\newcommand{\Rv}[1]{\ensuremath{\mathbb{R}^{#1}}}				% set of real-valued vectors
\newcommand{\R}[2]{\ensuremath{\mathbb{R}^{#1\times #2}}}		% set of real-valued matrices
\usepackage{xspace}% http://ctan.org/pkg/xspace
\algnewcommand{\algorithmicgoto}{\textbf{go to}}%
\algnewcommand{\Goto}{\algorithmicgoto\xspace}%
\algnewcommand{\Label}{\State\unskip}

\hypersetup{pdfinfo={
   Author={Andrea Del Prete, Francesco Romano et al.},
   Title={Prioritized Optimal Control},
   CreationDate={D:20130915},
   Subject={Robots},
   Keywords={Robots;Control},
}}

\title{\LARGE \bf
Prioritized Optimal Control
}

\author{Andrea Del Prete$^{1}$, Francesco Romano$^{2}$, Lorenzo Natale$^{1}$, Giorgio Metta$^{1}$, Giulio Sandini$^{2}$, Francesco Nori$^{2}$% <-this % stops a space
\thanks{*This paper was supported by the FP7 EU projects CoDyCo (No. 600716 ICT 2011.2.1 Cognitive Systems and Robotics), and Koroibot (No. 611909 ICT-2013.2.1 Cognitive Systems and Robotics).}%
\thanks{$^{1}$Del Prete, Metta and Natale are with the iCub Facility, Istituto Italiano di Tecnologia, Genova, Italy. %
        Email: {\tt\small name.surname@iit.it}}%
\thanks{$^{2}$Romano, Nori and Sandini are with the RBCS Department, Istituto \mbox{Italiano} di Tecnologia, Genova, Italy. %
        Email: {\tt\small name.surname@iit.it}}%
}

\begin{document}

\maketitle
\thispagestyle{empty}
\pagestyle{empty}

%%%%%%%%%%%%%%%%%%%%%%%%%%%%%%%%%%%%%%%%%%%%%%%%%%%%%%%%%%%%%%%%%%%%%%%%%%%%%%%%
\begin{abstract}
This paper presents a new technique to control highly redundant mechanical systems, such as humanoid robots.
We take inspiration from two approaches. 
\emph{Prioritized control} is a widespread multi-task technique in robotics and animation: 
tasks have strict priorities and they are satisfied only as long as they do not conflict with any higher-priority task.
\emph{Optimal control} instead formulates an optimization problem whose solution is either a feedback control policy or a feedforward trajectory of control inputs.
We introduce strict priorities in multi-task optimal control problems, as an alternative to weighting task errors proportionally to their importance.
This ensures the respect of the specified priorities, while avoiding numerical conditioning issues.
We compared our approach with both prioritized control and optimal control with tests on a simulated robot with 11 degrees of freedom.
\end{abstract}
%%%%%%%%%%%%%%%%%%%%%%%%%%%%%%%%%%%%%%%%%%%%%%%%%%%%%%%%%%%%%%%%%%%%%%%%%%%%%%%%

%!TEX root =  ../PrioritizedOC.tex
\section{Introduction}
Control of complex mechanical systems such as humanoids and legged robots is still a main concern for the control community, mainly because they are highly nonlinear and underactuated.
Given the high number of degrees of freedom (DoFs), it is often natural to formulate the control objective in terms of multiple tasks to achieve at the same time.
For instance, to make a humanoid robot walk, we have to control the motion of its center of mass and its swinging foot, the force exerted by its supporting foot and its whole-body posture.
In case of conflict between two or more tasks, we may desire that the most important task is achieved, at the expenses of the others.
This control approach --- known as prioritized or hierarchical control --- has been used in robotics and computer animation since the 80's \cite{Nakamura1987}.
Researchers have applied prioritized control in different forms.
Nakamura et al. \cite{Nakamura1987} and Siciliano et al. \cite{Siciliano1991} used it to control the cartesian velocity of multiple points of a robotic structure.
Sentis \cite{Sentis2008} was the first one to apply this technique on the robot dynamics, extending the Operational Space formulation \cite{Khatib1987}.
This allowed the control of contact forces, besides cartesian and joint-space motion.
In recent years, new formulations \cite{Righetti2011, Mistry2011} have contributed to improving the computational efficiency of this approach.
Mansard, Escande and Saab\cite{Saab2011, Escande2014} have studied efficient ways to include inequalities in the problem formulation, 
which can be used to model joint limits and motor torque bounds.
Research in computer animation \cite{DeLasa2009, DeLasa2010} has followed a similar path, trying to generate artificial motion by solving one or more Quadratic Programming (QP) problems (i.e. optimization problems with quadratic costs and linear constraints).

Another well-known technique is \emph{optimal control}, which specifies the control objective as the sum of a running cost (e.g. an integral through time of the tracking error and the control effort) and a terminal cost \cite{Kirk1970}. 
The result of the optimization is either a feedback control policy or a feedforward control input trajectory.
Of particular interest for robotics is \emph{Model Predictive Control} (MPC), also known as receding horizon control \cite{Diehl2009,Dimitrov2011,Tassa2012}. 
This class of optimal control methods computes at each time step the optimal control trajectory over a finite time horizon, but applies only the first value of the trajectory.
At the next time step the horizon shifts forward one time step and the process is repeated.
In case of multiple tasks, task errors are usually weighted proportionally to their priorities.
This can be problematic because too-low weights may lead to interferences between tasks, whereas too-large weights may introduce numerical conditioning issues.

Prioritized control and optimal control are strictly related: the first can indeed be formulated as a cascade of instantaneous optimal control problems \cite{Peters2007}.
Prioritized control can accomplish complex behaviors, while being efficient and straightforward to implement.
However, the main issue of prioritized control is its \emph{blindness}, meaning that it does not take into account the future evolution of the state.
This makes difficult to manage kinematic singularities and it does not address the problem of planning reference trajectories.
Moreover, an often overlooked problem is that a low-priority task may lead the system into a state in which a high-priority task becomes unfeasible.
This happens because priorities are satisfied only \emph{instantaneously}, since the problem is formulated at velocity/acceleration level.

Conversely, optimal control takes decisions in accordance to future predictions, but it does not handle properly the multi-task scenario.
We introduce task prioritization in optimal control, resulting in an approach that possesses the benefits of both control strategies.
Starting from an initial control and state trajectory, we linearize the system dynamics and solve a cascade of QPs, guaranteeing the specified task priorities.
Doing so, we avoid the problem of hand-tuning the task weights, which is particularly important in case of a large number of tasks.

This paper is structured as follows.
Section~\ref{sec:method} describes our theoretical results.
Section~\ref{sec:tests} reports two tests that compare our algorithm to both prioritized control and optimal control.
Finally, Section~\ref{sec:conclusions} draws the conclusions and discusses the future work.

%iTaSC \cite{Decre2013}, iLQG \cite{Todorov2005}, iLQR \cite{Li2003}.
%MPC for humanoid walking \cite{Dimitrov2011}.
%Whole-body MPC \cite{Tassa2012}, 7 times slower than realtime for humanoids.
%!TEX root =  ../PrioritizedOC.tex
\section{Method}
\label{sec:method}
This section illustrates the proposed control approach, which extends classic optimal control by introducing strict priorities between tasks.
Consider the dynamics of a nonlinear discrete system, with state $x \in \Rv{n}$, and control input $u \in \Rv{m}$, over a finite horizon $N$:
\begin{equation*}
x_{t+1} = f(x_t, u_t), \qquad t = 0, \dots, N-1
\end{equation*}
%We can obtain the discrete dynamics from the continuos dynamics by means of any discretization scheme (e.g. Euler).
In the following we denote with $X = \mat{x_1^\T \dots x_N^\T}^\T \in \Rv{nN}$ the entire state trajectory (apart from the initial state $x_0$) and with $U = \mat{u_0^\T \dots u_{N-1}^\T}^\T \in \Rv{mN}$ the entire control trajectory.
Assume that the system has to perform $K$ tasks, with task 1 having the highest priority, and task $K$ the lowest.
The objective of task $k$ is to minimize an arbitrary function of the state and control trajectories $g_k(X, U)$, while not interfering with any higher priority task:
\begin{equation} \label{eq:nonlin_prob}\begin{aligned}
g_k^* = &\min_{U\in \Rv{mN}} g_k(X,U) & \\ 
& \st \quad  x_{t+1} = f(x_t, u_t) \qquad & t = 0, \dots, N-1\\
& \qquad x_0 = x_s & \\
& \qquad g_i(X,U) = g_i^*            \qquad & \forall i<k \text{,}
\end{aligned} \end{equation}
where $x_s \in \Rv{n}$ is the initial state.
We take inspiration from Newton's method for nonlinear minimizations.
The key idea is to start from some nominal state and control trajectories and to solve the problem iteratively.
At each iteration we build linear-quadratic minimizations that describe the effects of small perturbations of $x$ and $u$ on the system dynamics and on the cost functions.
In particular, around the current solution, we take a linear approximation of the system dynamics and a quadratic approximation of the cost function and the priority constraints.
These ideas are common in the optimal control community.
For instance, Differential Dynamic Programming \cite{Murray1984} and the closely-related iterative LQR (iLQR) \cite{Li2003} are based on the same principles.

\subsection{Primary Task Resolution}
To start the optimization we need an initial guess for the control trajectory $\bar{U} \in \Rv{mN}$ (in case one has no insight on the solution he can set it to zero).
Setting $U=\bar{U}$ we simulate the system to get the state trajectory $\bar{X}$.
Consider small variations of the state and the control ($\dx{t}$ and $\du{t}$) around the current trajectories.
We can approximate the effect of these variations with a first-order Taylor expansion of the system dynamics:
\begin{align*}
\bar{x}_{t+1} + \dx{t+1} &\simeq f(\bar{x}_t, \bar{u}_t) + A_t \dx{t} + B_t \du{t} \\
\dx{t+1} &\simeq A_t \dx{t} + B_t \du{t},
\end{align*}
where $A_t = \pder{f(\bar{x}_t,\bar{u}_t)}{x} \in \R{n}{n}$ and $B_t = \pder{f(\bar{x}_t,\bar{u}_t)}{u} \in \R{n}{m}$.
Since $\dx{0}=0$, we can describe the whole evolution of the variational state in a single equation:
\begin{equation*}
\DX = G \DU,
\end{equation*}
where $\DX = \mat{\dx{1}^\T \dots \dx{N}^\T}^\T \in \Rv{nN}$ is the entire variational state trajectory, $\DU = \mat{\du{0}^\T \dots \du{N-1}^\T}^\T \in \Rv{mN}$ is the entire variational control trajectory, and $G \in \R{nN}{mN}$ is the matrix mapping the control trajectory to the state trajectory.
To simplify the derivation we assume that all the costs take this form:
\begin{equation*}
g_k(X, U) = \half || c_k(X) ||^2 + \half U^\T E_k U,
\end{equation*}
where $E_k \in \R{mN}{mN}$ is a positive-semidefinite matrix, i.e. $E_k\succeq 0$.
This form is general enough for robotics applications: the state dependent cost $||c_k(X)||^2$ can represent any task-space tracking/reaching error, whereas the \mbox{control-dependent} cost $U^\T E_k U$ can penalize some form of effort. In practice, since the effort minimization consumes all the redundancy, $E_k$ must be zero for all tasks but the last one.
If needed, it is possible to extend the following derivation for a more general cost function.

Let us consider the effect that state and control variations have on the cost of task $k$:
\begin{align*}
\delta g_k(\DX,\DU) =& g_k(\bar{X}+\DX, \bar{U}+\DU) - g_k(\bar{X}, \bar{U})
\end{align*}
For small variations, we can approximate $\delta g_k$ with $\delta \tilde{g}_k$ by replacing $g_k(X,U)$ with its second-order Taylor expansion in $X=\bar{X}$ and $U = \bar{U}$:
\begin{equation} \label{eq:deltag_approx} \begin{aligned}
\delta \tilde{g}_k{=}& c_k(\bar{X})^\T C_k \DX + \bar{U}^\T E_k \DU + \tfrac{1}{2} \DU^\T E_k \DU \\
& + \tfrac{1}{2} \DX^\T( C_k^\T C_k + c_k(\bar{X})^\T \pder{C_k}{X}) \DX,
\end{aligned} \end{equation}
where $C_k = \pder{c_k}{X}(\bar{X})$.
We can now formulate an optimization problem to find the control variations that result in the minimum cost variation, subject to the variational dynamics. 
In particular, for task $k=1$ we have:
\begin{equation} \label{eq:lin_prob}\begin{aligned}
\delta g_1^* = &\min_{\DU\in \Rv{mN}} \delta \tilde{g}_1(\DX,\DU) + s_1||\DX||^2 + r_1 ||\DU||^2 \\ %p_k(\DX, \DU) \\ 
& \st \quad  \DX = G \DU,
\end{aligned} \end{equation}
where $s_1 \in \mathbb{R}$ and $r_1 \in \mathbb{R}$ are strictly positive weights that penalize deviations from the current nominal solution $(\bar{X},\bar{U})$.
These penalties are fundamental because the linear approximation of the system dynamics is valid only in the vicinity of the current solution.
Since task 1 has the highest priority, this problem does not contain any constraint besides the system dynamics.
To solve \eqref{eq:lin_prob} we substitute the constraint into the cost function, and we set the derivative of the cost function with respect to $\DU$ equal to zero.
Doing so we get:
\begin{align*}
S_1 \DU &= d_1,
\end{align*}
where:
\begin{equation} \label{eq:def_SHd} \begin{aligned}
S_1 &= G^\T H_1 G + s_1 G^\T G + E_1 + r_1 I \\
H_1 &= C_1^\T C_1 + c_1(\bar{X})^\T \pder{C_1}{X} \\
d_1 &= -G^\T C_1^\T c_1(\bar{X}) - E_1 \bar{U}
\end{aligned} \end{equation}
Since $r_1 > 0$ and $E_1\succeq 0$, this system of equations has a unique solution, that is:
\begin{align*}
\DU_1^* &= S_1^{-1} d_1
\end{align*}
At this point we simulate the system using the new control trajectory $\bar{U}+\DU_1^*$, we obtain the new state trajectory \mbox{$\bar{X}+\DX_1^*$}, and we check whether the cost of the task has improved:
\begin{equation}\label{eq:cost_impr}
g_1(\bar{X}+\DX_1^*, \bar{U}+\DU_1^*) < g_1(\bar{X}, \bar{U})
\end{equation}
If the cost has not improved, it means that we have moved into a region of the state/control space in which our approximations are no longer accurate.
To improve the solution we iteratively increase the penalty weights $(s_1, r_1)$ and recompute the solution of \eqref{eq:lin_prob}, until \eqref{eq:cost_impr} is satisfied.
Once we have found a $\DU_1^*$ that gives an improvement of the cost, we move to the following tasks.

\subsection{Cost Constraint Linearization}
Starting from the second task (i.e. $\forall k>1$), the minimization is subject to the priority constraints, which impose not to alter the cost of the tasks with higher priority.
In terms of variations we can express these constraints as:
\begin{equation} \label{eq:cost_constr}
\delta g_i(\DX,\DU) = \delta g_i^* \qquad \forall i<k
\end{equation}
In general, nonlinear equality constraints needs to be approximated with a linear function in order to find analytical expressions of their solutions (as we did for the system dynamics).
However, these nonlinear constraints have a special form that allows us to use quadratic approximations, which are more accurate.
We exploit the fact that a convex quadratic function is equal to its minimum value if and only if its gradient is zero.
This simple observation allows us to convert the quadratic approximations of the cost constraints \eqref{eq:cost_constr} into linear constraints:
\begin{equation*}
\pder{\delta \tilde{g}_i(\DX,\DU)}{\DU} = 0 \qquad \forall i<k,
\end{equation*}
where $\delta \tilde{g}_i$ is the quadratic approximation of $\delta g_i$, as defined in \eqref{eq:deltag_approx}.
Computing this derivative we get:
\begin{align} \label{eq:Dcost_constr}
\underbrace{(G^\T H_k G + E_k)}_{T_k} \DU &= d_k,
\end{align}
where $H_k$ and $d_k$ have already been defined in \eqref{eq:def_SHd}.
Typically we expect that, for all the tasks but the last one, the matrix $T_k \in \R{mN}{mN}$ has a nontrivial nullspace (because \mbox{$E_k=0$} $\forall k \ne K$).
If that is the case, the constraint \eqref{eq:Dcost_constr} admits infinite solutions:
\begin{align*} \label{eq:Dcost_constr}
\DU &= T_k^+ d_k + \underbrace{(I - T_k^+ T_k)}_{P_k} \DU_0,
\end{align*}
where $\DU_0 \in \Rv{mN}$ is an arbitrary vector and $T_k^+$ denotes the Moore-Penrose pseudoinverse of $T_k$.

\subsection{Secondary Tasks Resolutions}
We write the linearization of the problem $k$ as:
\begin{equation} \label{eq:lin_prob_sec}\begin{aligned}
\delta g_k^* = &\min_{\DU_0 \in \Rv{mN}} \delta \tilde{g}_k(\DX,\DU) + s_k||\DX||^2 + r_k ||\DU||^2 \\
& \st \quad  \DX = G \DU \\
& \qquad \DU = T_i^+ d_i + P_i \DU_0 \qquad \forall i<k
\end{aligned} \end{equation}
The problem consists in minimizing a quadratic function subject to linear equality constraints, so we can find its solution analytically.
We substitute the constraints inside the cost function and then we set its derivative with respect to $\DU_0$ to zero.
As for the primary task, after computing the solution we must check that the cost has actually improved.
Moreover, we check that the costs of the higher-priority tasks have not increased within a tolerance $\epsilon > 0$
 % (in our tests we used $\epsilon = 10^{-6}$)
.
If the cost of either the current task or a higher-priority task has increased, then we must increase the penalty terms $(s_k, r_k)$ and repeat the optimization.
To solve the whole cascade of minimizations we used Algorithm \ref{alg:task_lin}.
\begin{algorithm} 
\caption{Resolution of the Linearized Hierarchy of Tasks.} \label{alg:task_lin}
\begin{algorithmic}[1]
	\Procedure{SolverStep}{$\bar{X}, \bar{U}, G, \{c\}, \{E\},\{r\},\{s\}$}
	\State $\DU \gets 0$	\Comment{Initialize variational control trajectory}
	\State $P \gets I$     \Comment{Initialize nullspace projector}
	\For{$\text{task } k = 1 \to K$} 
		\State $H \gets C_k^\T C_k + c_k(\bar{X})^\T \pder{C_k}{X}$
		\State $T \gets G^\T H G + E_k$
		\State $d \gets -G^\T C_k^\T c_k(\bar{X}) - E_k \bar{U}$
		\Label \texttt{WeightTuningLoop:}
		\State $S \gets T + s_k G^\T G + r_k I$ \label{marker}
		\State $\DU \gets \DU + (SP)^+(d - S \DU)$
		\For{$\text{task } i = 1 \to k$} 
			\State $\delta g_i \gets $ \text{computeCostVariation($i, \bar{X},\bar{U},\DU$)}
			\If{$\delta g_i > \epsilon$}
				\State $(s_k,r_k) \gets$ \text{updatePenalties($s_k,r_k$)}
				\State \Goto \texttt{WeightTuningLoop}
			\EndIf
		\EndFor
		\State $P \gets P - (TP)^+ TP$
	\EndFor
	\State \Return $\DU$
	\EndProcedure
\end{algorithmic}
\end{algorithm}

\subsection{Penalty Weight Tuning}
The tuning of the penalty weights $(s_k,r_k)$ is crucial for the fast convergence of the algorithm.
%In Algorithm \ref{alg:task_lin} we did not describe completely our policy for tuning the penalty weights, which we report here.
Ideally, at each iteration one would like to find the values of $(s_k,r_k)$ that result in the biggest reduction of the cost, but this search can be expensive.
We can summarize our policy for tuning the penalty weights as follows.
If the cost increases we multiply $(s_k,r_k)$ times a factor $\mu > 1$
 % (we used $\mu = 1.5$)
  and repeat the optimization until we get an improvement in the cost.
If the cost decreases we divide $(s_k,r_k)$ by $\mu$ and recompute the solution.
As long as the cost decreases, we continue decreasing $(s_k,r_k)$, and we stop only when there is an increase in the cost.
At each iteration we initialize $(s_k,r_k)$ with the values that we found in the previous iteration.

\subsection{Convergence}
\label{subsec:convergence}
We test convergence at the end of each main iteration, i.e. after each execution of Algorithm \ref{alg:task_lin}.
After the iteration $i$, we have computed the state trajectory $X_i$ and the control trajectory $U_i$.
At this point, the algorithm stops if all the tasks have converged.
A task has converged if at least one of these values is below the associated threshold:
\begin{itemize}
    \item the absolute value of the cost $g_k(X_i,U_i)$;
    \item the relative cost improvement $\frac{|g_k(X_i,U_i) - g_k(X_{i-1},U_{i-1})|}{|g_k(X_{i-1},U_{i-1})|}$
\end{itemize}
% In particular, we set the threshold for the absolute cost to $10^{-6}$ and the threshold for the relative cost improvement to $10^{-2}$.

\subsection{Algorithm summary}
We here summarize the Prioritized Optimal Control algorithm sketched in Algorithm \ref{alg:main_alg}.
The input parameters are the initial state $x_s$, an initial control trajectory $U_0$, the state dependent costs $\{c\}$ and the effort weights $\{E\}$.
The initialization phase consists in initializing the penalty weights (used by the \emph{SolverStep} procedure described in Algorithm \ref{alg:task_lin}) 
and computing the initial state trajectory with the associated costs.
Then the main loop begins.
Each iteration starts by computing the mapping matrix $G$ and linearizing the nonlinear dynamics around the nominal trajectory $(X,U)$, thus obtaining a \mbox{time-varying} linear dynamics.
Then the variational control trajectory are computed by using Algorithm \ref{alg:task_lin}.
At this point the algorithm updates control and state trajectories ($U$, $X$) and the associated costs.
Finally the stopping criteria described in Section~\ref{subsec:convergence} are tested. 
If the criteria are met then the algorithm exits and returns the computed control trajectory, otherwise it updates the costs and iterates the procedure.

\begin{algorithm} 
\caption{Prioritized Optimal Control main iteration.} \label{alg:main_alg}
\begin{algorithmic}[1]
	\Procedure{PrioritizedOC}{$x_s, U_0, \{c\}, \{E\}$}

    \State $\{r\}, \{s\} \gets \text{initializeWeights()}$
	\State $U \gets U_0$	%\Comment{Initialize control law}
    \State $X \gets \text{simulateTrajectory}(x_s, U)$
    \State $\{J\} \gets \text{computeCosts}(X, U, \{c\}, \{E\})$

    \Loop
        \State $\{A_i\}, \{B_i\}, \{C_i\} \gets \text{linearizeSystem(X,U)}$
        \State $G \gets \text{computeG}(\{A_i\}, \{B_i\}, \{C_i\})$
        
        \State $\DU \gets \text{SolverStep}(X, U, G, \{c\}, \{E\},\{r\},\{s\})$
        
        \State $U \gets U + \DU$
        \State $X \gets \text{simulateTrajectory}(x_s, U)$
        \State $\{J_{new}\} \gets \text{computeCosts}(X, U, \{c\}, \{E\})$
        \If {stopCriterion($\{J\}, \{J_{new}\}$)}
            \State Break
        \EndIf
        \State $\{J\} \gets \{J_{new}\}$
    \EndLoop
    \State \Return $U, \{J\}$
	\EndProcedure
\end{algorithmic}
\end{algorithm}

%!TEX root =  ../PrioritizedOC.tex
%
% Ideas for tests:
% \begin{itemize}
% \item comparison with iLQR with task weighting
% \item add effort minimization task
% \item comparison with prioritized computed torque control (PCTC)
% \begin{itemize}
% \item show that near singularities it has problems if no damping factor is used
% \item show that it does not minimize effort
% \end{itemize}
% \item increase DoFs of puma (at least 12 DoFs) and increase number of tasks to show that, for many conflicting tasks, it's difficult to find the right weights (pass to robotran for doing this)
% \item make the robot underactuated removing one motor (i.e. zero torque at one joint)
% \end{itemize}
\section{Tests}
\label{sec:tests}
We tested our algorithm against both prioritized control and classic optimal control, on the same test case.
Our method generates open-loop trajectories, which cannot be applied without a stabilizer.
In the first test, we compared the open-loop trajectories generated by our method with those generated by a classic optimal control approach using task weighting.
In the second test, we used a prioritized controller to stabilize the trajectories computed by our method.
We tried to achieve the same tasks using approximately minimum-jerk trajectories and we compared the results.

The test platform was the upper body of a simulated humanoid robot, which had to reach three cartesian points with three parts of its body, while minimizing the effort. 
In order of decreasing priority, we controlled the position of these three points: left-arm end-effector, right-arm end-effector and the top of the torso.
The three tasks were individually feasible, but impossible to achieve at the same time.
This allowed us to check whether the task priorities were satisfied: we expected the left end-effector to perfectly reach its target, while we expected significant errors on the right end-effector and the top torso.

\begin{figure}[tbp]
\centering
\subfloat[0 s]{\includegraphics[height=2.9cm]{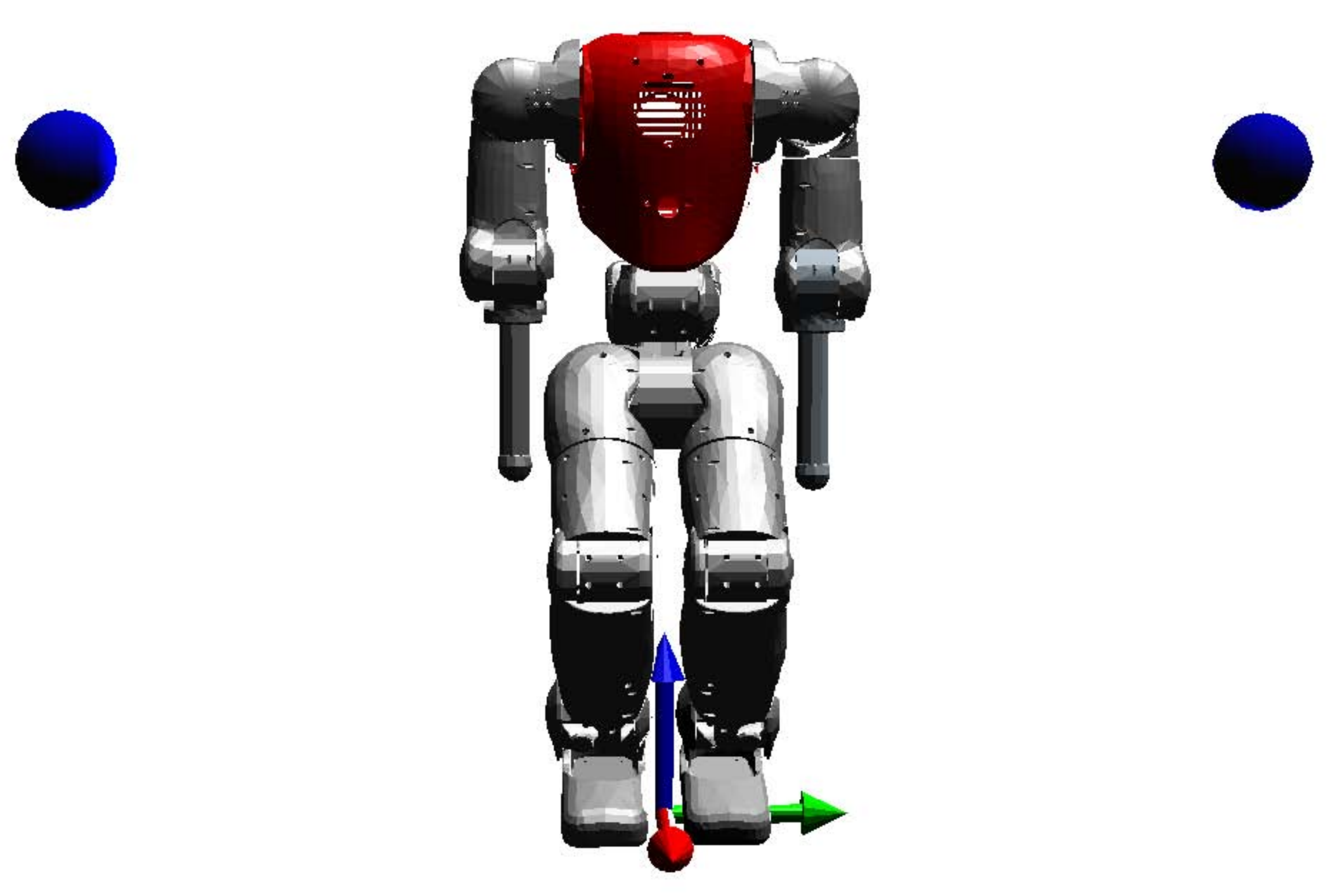}} 
\subfloat[0.5 s]{\includegraphics[height=2.9cm]{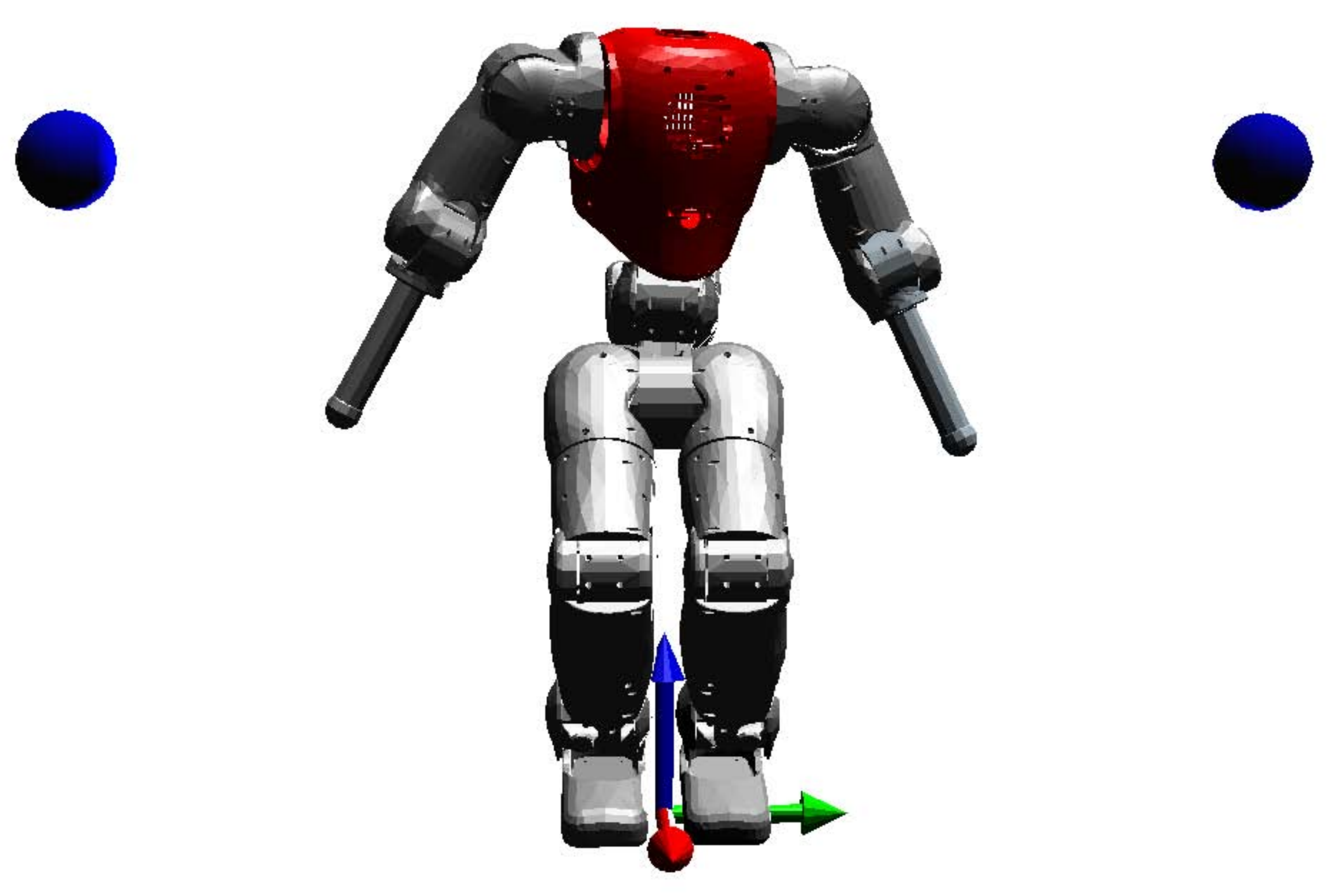}} \\
\subfloat[0.9 s]{\includegraphics[height=2.9cm]{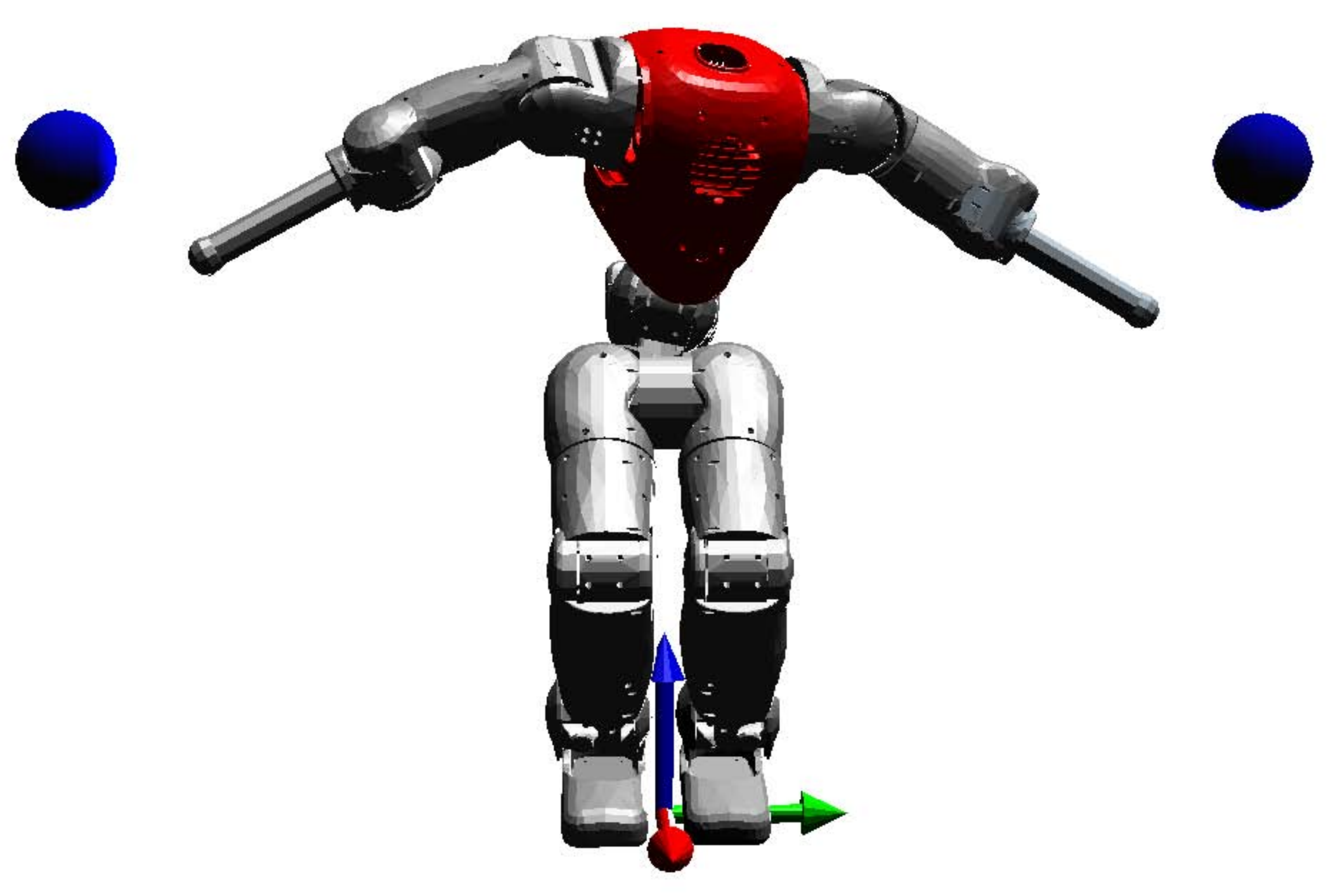}} 
\subfloat[1.1 s]{\includegraphics[height=2.9cm]{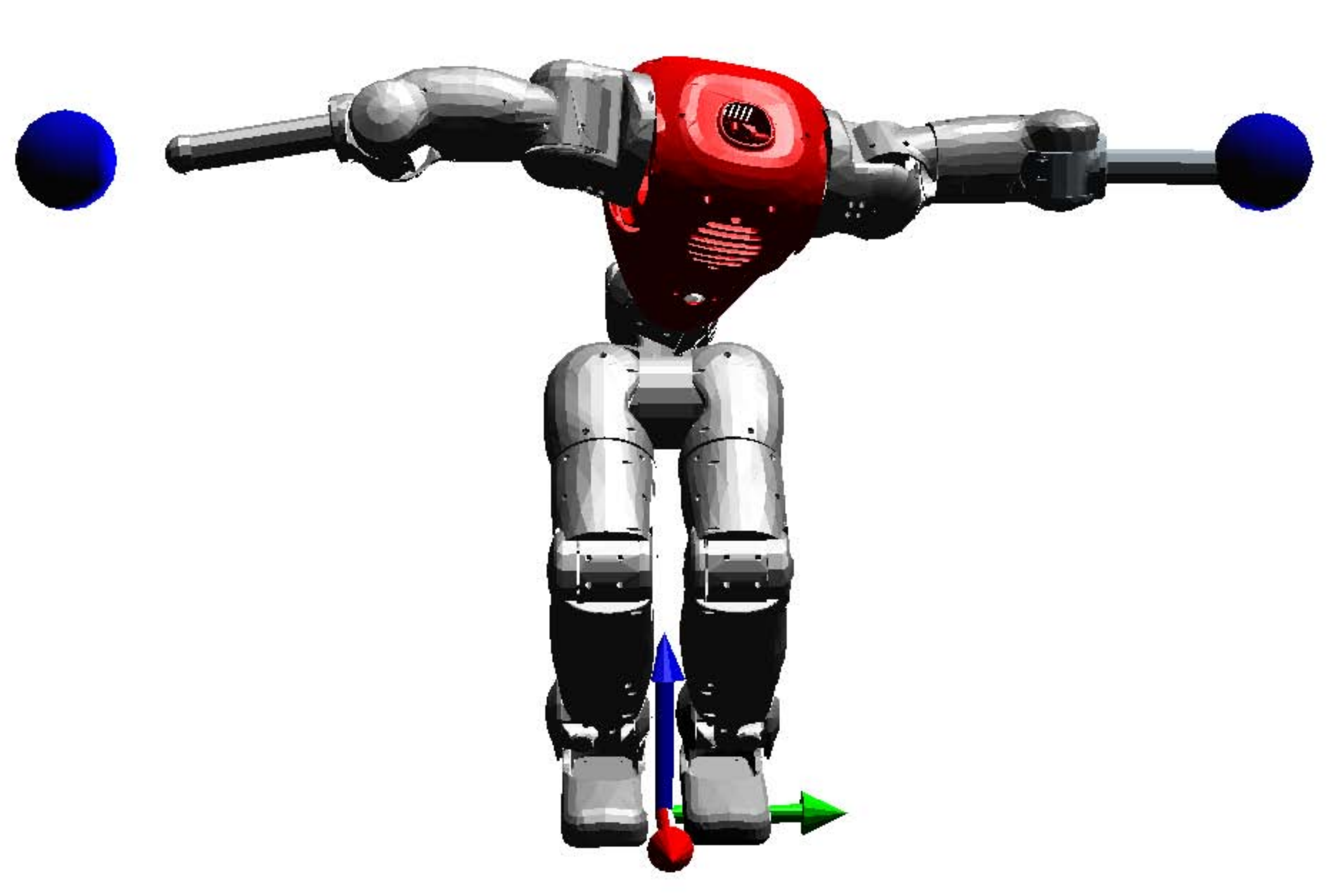}} \\
\caption{Screenshots of the simulation. The two blue balls represent the targets of the left (high priority) and right (low priority) end-effectors of the robot. }
\label{fig:screenshots}
\end{figure}

We discretized the equations of motion of the system by using a first-order explicit Euler scheme with a fixed time step of 20 ms.
We set the time horizon to 1 s, which resulted in a total of $N=50$ time steps.
In all tests, we initialized the control trajectory with gravity compensation torques, so that the robot maintained the initial state for the whole time horizon.
Fig.~\ref{fig:screenshots} reports some screenshots of the simulation, showing also two reference points.

We tested our approach on a customized version of the Compliant huManoid (CoMan) simulator \cite{Dallali}.
The base of the robot was fixed because we used only its upper body, which counts 11 DoFs: 4 in each arm and 3 in the torso.
We integrated the equations of motion with the Simulink variable step integrator \emph{ode23t} (relative tolerance $10^{-3}$, absolute tolerance $10^{-6}$).
Table~\ref{tab:hyperparameters} lists the values used during our tests for all the parameters of the algorithm.
The \emph{pseudoinverse tolerance} is used to discriminate zero and nonzero singular values when computing pseuoinverses.
%As a convergence criterion we stopped the algorithm when either the cost becomes smaller than $10^{-6}$ or the relative cost improvement between two iterations is less than $1\%$.

\subsection{Comparison with Classic Optimal Control}
\label{sec:test1}
In this test we compared the performances of our algorithm with those of classic optimal control. 
To approximate strict priorities with classic optimal control we minimize the following cost function:
\begin{equation}
    g(X,U) = w^2 g_1(X,U) + w g_2(X,U) + g_3(X,U) \text{,}
\end{equation}
where $w \in \Rv{}$ is a user-selected weight parameter.
Table~\ref{tab:weigths_comparison} shows the results obtained with different values of $w$.
By increasing $w$, the error of the primary task got lower, while the errors of the other tasks became higher.
However, increasing $w$ did not yield a clear trend on the secondary tasks. 
This outlines the difficulty of tuning the task weights. 
We also verified that, whenever we changed the tasks, we had to repeat the weight-tuning procedure.

During the tests, we noticed different problems as we incremented $w$. 
First of all, the number of iterations needed to converge increased (see Table \ref{tab:weigths_comparison}).
Second, we had to increase the initial value of the penalty weights ($s$, $r$) in the linear approximation of the system. 
Failing to do that resulted in an early termination of the optimization problem for numerical reasons.
Third, we had to adapt the tolerances used in the convergence criteria. 
Since the cost function is a weighted sum of the task errors, its value has not a consistent unit of measurement.

Conversely, our method managed to obtain an acceptable error on the first task ($0.5$ mm), while obtaining significantly smaller errors on the second and third tasks.
The number of iterations needed to converge was lower, but one iteration of our algorithm is in general more expensive, so this comparison is not fair.
\begin{table}[t]%[ht]
\caption{Values of the parameters of the algorithm used in the tests.}
\label{tab:hyperparameters}
\centering 
\begin{tabular}{p{1.5cm} | p{1cm} || p{3.0cm} | p{1cm}} 
\hline 
% \rowcolor[gray]{.9}
    Parameter & Value & Parameter & Value \\
    \hline
    \rowcolor[gray]{.9} Initial $s_{k}$            & 1                   & Absolute Cost Threshold     & $10^{-6}$ \\
                                  Initial  $r_{k}$            & $10^{-2}$     &Relative Cost Threshold       & $10^{-2}$     \\
    \rowcolor[gray]{.9} $\mu$                        & 1.5               & Pseudoinverse Tolerance & $10^{-5}$ \\
                                  $\epsilon$                 & $10^{-6}$     &  & \\
\hline 
\end{tabular}
\end{table}
\begin{table}[t]%[ht]
\caption{Comparison between classic Optimal Control and Prioritized Optimal Control. Each column shows error norm for three different choices of the weight $w$ and the number of iterations needed to compute the solution. The last column shows the same results for our algorithm.}
\label{tab:weigths_comparison}
\centering 
\begin{tabular}{p{1.9cm} | p{1.1cm} p{1.1cm} p{1.1cm}| p{1.1cm} } 
\hline 
       	& \multicolumn{3}{ c |}{Classic Optimal Control}	& Prioritized\\
                  & \multicolumn{1}{ c}{$w=10^2$} & \multicolumn{1}{ c}{$w=10^3$} & \multicolumn{1}{ c|}{$w=10^4$} & \multicolumn{1}{ c}{O.C.} \\
	 [0.5ex] \hline \rowcolor[gray]{.9}
	 Task 1 Err. [\SI{}{\micro\metre}]&	$530$			& $54$ 				&  $27$  			&   $507$ 	\\ 
	 Task 2 Err. [mm]&	        $142$			&  $307$   			&  $290$  			&  $107$    			\\ \rowcolor[gray]{.9}
	 Task 3 Err. [mm]&	        $359$			& $426$				&  $425$ 			&   $305$ 	\\
%\begin{tabular}{p{1.3cm} | p{1.4cm} p{1.3cm} p{1.3cm}| p{1.2cm} } 
%\hline 
%       	& \multicolumn{3}{ c |}{Classic Optimal Control}	& Prioritized\\
%                  & \multicolumn{1}{ c}{$w=10^2$} & \multicolumn{1}{ c}{$w=10^3$} & \multicolumn{1}{ c|}{$w=10^4$} & \multicolumn{1}{ c}{O.C.} \\
%	 [0.5ex] \hline \rowcolor[gray]{.9}
%	 Task 1	Err.&	$530$ $10^{-3}$	& $54$ $10^{-3}$		&  $27$ $10^{-3}$ 	&   $507$ $10^{-3}$ 	\\ 
%	 Task 2 Err.&	$142$			&  $307$   			&  $290$  			&  $107$    			\\ \rowcolor[gray]{.9}
%	 Task 3 Err.&	$359$			& $426$				&  $425$ 			&   $305$ 	\\
     \hline 
	 Iterations &	    $22$     	    	& $54$ 				&         $75$    			&   	$18$		 \\ 
[0.5ex] \hline 
\end{tabular}
\end{table}

\subsection{Comparison with Prioritized Control}
\label{sec:test2}
In this test we used an implementation of the prioritized control formulation presented in \cite{DelPrete2014a}.
To prevent deviations from the desired trajectory and to ensure disturbance rejection, the controller computed the desired task-space accelerations $\ddot{x}^* \in \Rv{3}$ with a proportional-derivative feedback control law:
\begin{equation*}
\ddot{x}^* = \ddot{x}_r + K_d (\dot{x}_r - \dot{x}) + K_p (x_r - x),
\end{equation*}
where $x_r(t), \dot{x}_r(t), \ddot{x}_r(t) \in \Rv{3}$ are the position-velocity-acceleration reference trajectories, whereas $K_d \in \R{3}{3}$ and $K_p\in \R{3}{3}$ are diagonal matrices acting as derivative and proportional gains, respectively.
We set all the diagonal entries of $K_p$ to $10 s^{-2}$, and all the diagonal entries of $K_d$ to $5 s^{-1}$.
The controller also includes a simple trajectory generator \cite{Pattacini2010}, which provides approximately-minimum-jerk trajectories.

Prioritized control tends to make robots unstable when trying to move close to a singularity.
A common approach to avoid instability is to use damped pseudoinverses \cite{Chiaverini1994}, which however degrade tracking performances.
In this test the robot reached a singular configuration, because it had to completely stretch both arms to get as close as possible to the targets.
Using Moore-Penrose pseudoinverses resulted in instability, despite the low control period of 1 ms.
With prioritized control, we had to set the damping factor to $0.1$ to avoid instability.
\begin{figure}[tbp]
\centering
\subfloat[Prioritized control. The reference trajectory is a quasi-minimum jerk trajectory, designed to reach the desired target in about 1 s.]{\includegraphics[width=0.48\textwidth]{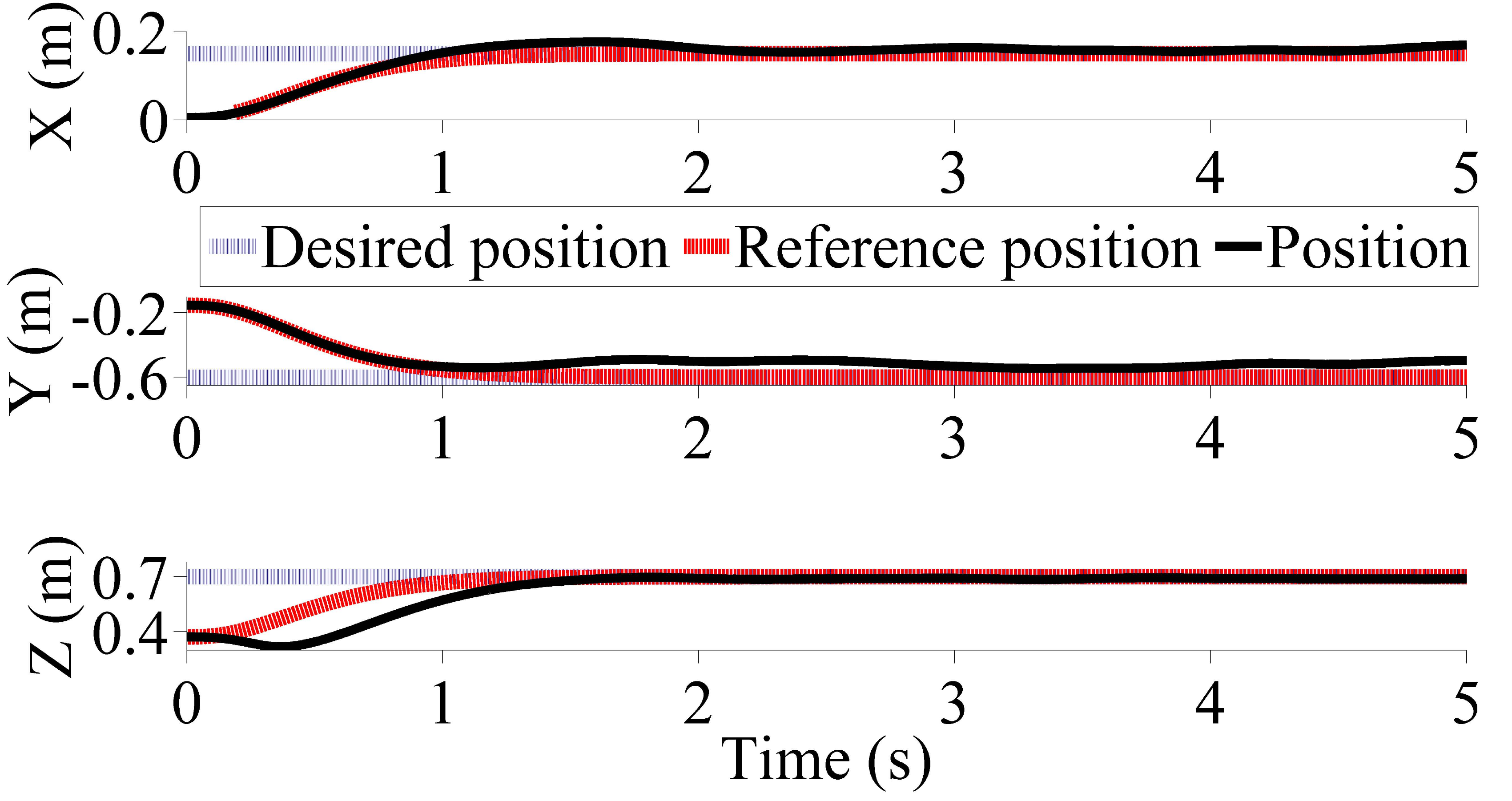}} \\
\subfloat[Prioritized control + Prioritized optimal control. The reference trajectory is found by our algorithm with a time horizon of 1 s.]{\includegraphics[width=0.48\textwidth]{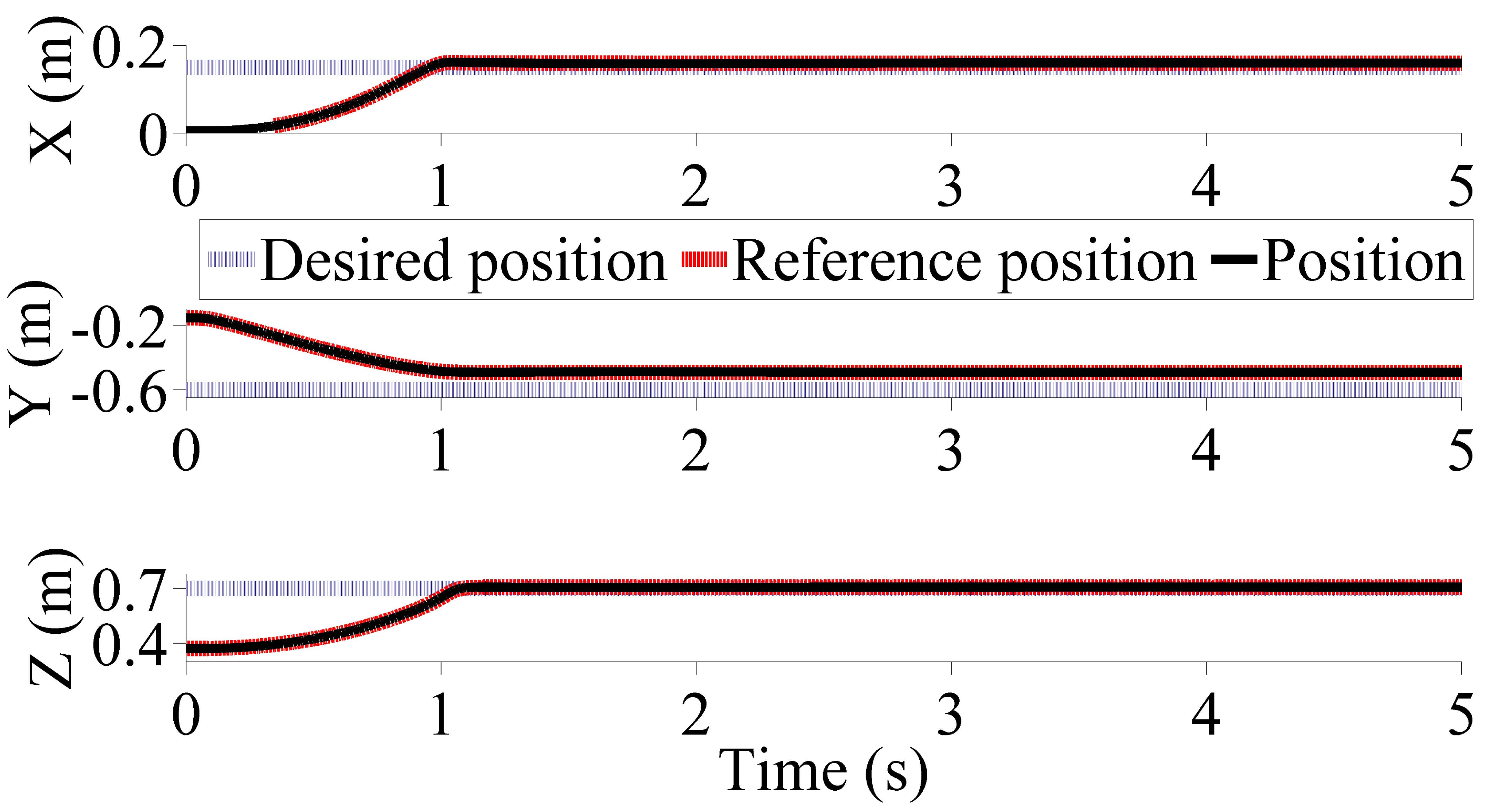}}
\caption{Test 2. Comparison of right hand reaching between prioritized control (a) and ``prioritized control + prioritized optimal control'' (b).}
\label{fig:test2}
\end{figure}
Despite the high damping factor (typically we set it to $0.02$), the robot exhibited an oscillatory behavior around the goal configuration. 
Fig.~\ref{fig:test2} depicts the trajectory of the right hand of the robot, clearly showing the oscillations.
The left end-effector manifested a similar behavior.

Our algorithm generates open-loop trajectories, which cannot be used without a stabilizer.
We decided to use the same prioritized controller as stabilizer for the trajectories computed by our method.
In this case we could safely lower the damping factor of the pseudoinverses to $0.01$, because the reference trajectories were always feasible.
The improvement in the resulting behavior was remarkable: the controller was able to track the trajectories with negligible errors.
The robot reached the goal configuration and Fig.~\ref{fig:test2} shows no oscillation around the final positions for the right end-effector.
Also the left end-effector and the top of the torso were stable. \addtolength{\textheight}{-0.5 cm}
%!TEX root =  ../PrioritizedOC.tex
\section{Conclusions}
\label{sec:conclusions}
We presented an optimal control algorithm for prioritized multi-task scenarios.
Control of multiple tasks with strict priorities is common in robotics and computer animation, especially with high-DoF systems.
Classical optimal control can tackle multi-task scenarios by minimizing a weighted sum of the task objectives, in which weights are proportional to the task priority levels.
However, as we showed in Section~\ref{sec:test1}, improper weights may lead to either violations of priorities or suboptimal solutions.
Conversely, our method guarantees the respect of task priorities and it is free of numerical conditioning issues.
At the same time, it spares the user the tuning of task weights.
We also demonstrated the benefits of our approach with respect to prioritized control.
We computed reference trajectories using prioritized optimal control and we fed them to a prioritized controller.
In this way, we managed to eliminate the oscillations due to the vicinity to a singular configuration.

Our final goal is to apply prioritized optimal control on a real humanoid robot, but before doing this we need to address some issues.
Humanoids are often in contact with the environment, but we did not consider contact forces in our derivation.
We plan to extend our algorithm to model and control interaction forces.
Accounting for joint limits and motor torque bounds is also fundamental for real-life applications.
By adding inequality constraints inside the control problem we could model these bounds.
Finally, we believe that computation time is the main obstacle preventing the implementation of online trajectory optimization on humanoid robots \cite{Tassa2012}.
Our intention is to code our algorithm in C++ (rather than in Matlab) and to provide an efficient open-source software library for model predictive control.
In case real-time performances were impossible to attain, we can still use the presented approach to plan multi-task trajectories.
As we showed in our second test (see Section~\ref{sec:test2}), we can then use these trajectories as reference inputs for a prioritized controller.
The combination of prioritized optimal control (for planning) and prioritized control (for stabilization) could be a powerful approach for achieving complex behaviors on humanoid robots.

\bibliographystyle{IEEEtran}
\bibliography{IEEEabrv,references}

\end{document}